\pdfoutput=1

\documentclass[11pt]{article}

\usepackage[]{acl}
\usepackage{graphicx}

\usepackage{times}
\usepackage{latexsym}

\usepackage[T1]{fontenc}

\usepackage[utf8]{inputenc}

\usepackage{microtype}

\usepackage{booktabs}
\usepackage{multirow}
\usepackage{graphicx}
\usepackage{enumitem}
\usepackage{hyperref}

\newlist{mylist}{enumerate}{1}%
\setlist[mylist]{label={(\arabic*)}}%

\title{X-PuDu at SemEval-2022 Task 6: Multilingual Learning for English and Arabic Sarcasm Detection}

\author{Yaqian Han, Yekun Chai, Shuohuan Wang, Yu Sun \\ 
    Baidu \\
\texttt{\{hanyaqian,chaiyekun,wangshuohuan,sunyu02\}@baidu.com}\\
\AND Hongyi Huang, Guanghao Chen, Yitong Xu, Yang Yang \\
        Shanghai Pudong Development Bank \\
        \texttt{\{huanghy6,chengh13,xuyt3,yangy103\}@spdb.com.cn}
}

\begin{document}
\maketitle
\begin{abstract}

Detecting sarcasm and verbal irony from people's subjective statements is crucial to understanding their intended meanings and real sentiments and positions in social scenarios. This paper describes the X-PuDu system that participated in SemEval-2022 Task 6, \textit{iSarcasmEval - Intended Sarcasm Detection in English and Arabic}, which aims at detecting intended sarcasm in various settings of natural language understanding. Our solution finetunes pre-trained language models, such as ERNIE-M and DeBERTa, under the multilingual settings to recognize the irony from Arabic and English texts. Our system ranked second out of 43, and ninth out of 32 in Task A: one-sentence detection in English and Arabic; fifth out of 22 in Task B: binary multi-label classification in English; first out of 16, and fifth out of 13 in Task C: sentence-pair detection in English and Arabic.


\end{abstract}

\section{Introduction}


Sarcasm is the use of language that typically signifies the opposite to mock or convey contempt. As a narrow research field in natural language processing (NLP), sarcasm detection is a particular case in the spectrum of sentiment analysis, with important implications for a slew of NLP tasks, such as sentiment analysis, opinion mining, author profiling, and harassment detection. In the textual data, these tonal and gestural clues like heaving tonal stress and rolling of the eyes are missing, making it more difficult for machines.



 
 The sarcastic intention of human annotators has potentially hindered the training and evaluation process in detecting the genuine emotions and positions of the natural language. Thus, this task~\citep{abufarha-etal-2022-semeval} adopted a novel data collection method \citep{Oprea2020iSarcasmAD}, where authors themselves label the training samples. For sarcastic texts, the authors also rephrase them into non-sarcastic ones. Then, linguistic experts further checked the scathing pieces and labeled them into sub-categories of sarcasm defined by (\citealp{Leggitt2000}): sarcasm, irony, satire, understatement, overstatement, and rhetorical question. 
 




This SemEval task requires the identification of sarcasm in either one sentence or sentence pairs in various language settings, which consists of three subtasks:
\begin{mylist}
    \item Task A (English and Arabic): Given a text, determine whether it is sarcastic or non-sarcastic;
    \item Task B (English only): A binary multi-label classification task. Given a text, determine which ironic speech category it belongs to, if any;
    \item Task C (English and Arabic): Given a sarcastic text and its non-sarcastic rephrase, i.e. two texts that convey the same meaning,  determine which is  the sarcastic one.
\end{mylist}

Our method employed various multilingual or mono-lingual pre-trained language models, such as ERNIE-M~\citep{ouyang2020ernie} and DeBERTa~\citep{he2021debertav3} to address each component of this task, with a bunch of fine-tuning and ensemble techniques. Our system finally achieved 
\begin{itemize}
    \item 2nd out of 43 and 9th out of 32 in English and Arabic subtasks in \textit{Task A};
    \item 5th out of 22 in \textit{Task B};
    \item 1st out of 16 and 5th out of 13 in English and Arabic subtasks in \textit{Task C}.
\end{itemize}


\begin{figure*}[t]
\centering
\includegraphics[width=0.9\textwidth]{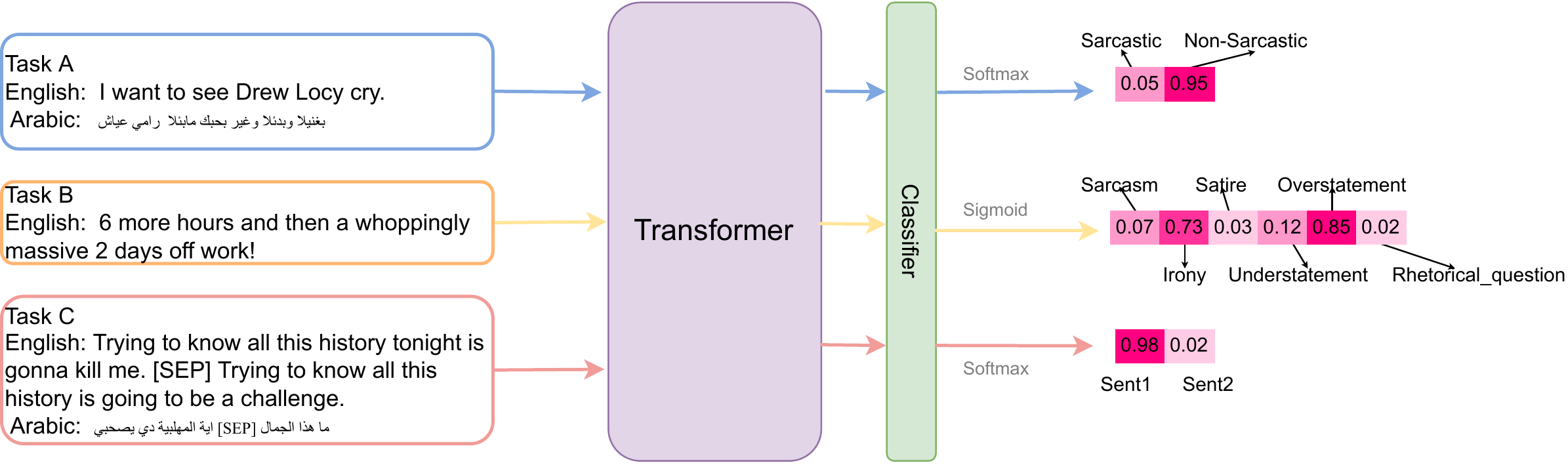}
\caption{Fine-tuning pre-trained models on the \textit{iSarcasmEval} data.}
\label{fig:arch}
\end{figure*}

\section{Previous Work}

After detecting sarcasm in the speech was firstly proposed in~\citep{Tepperman2006}, sarcasm detection has attracted extensive attention in the NLP community. Afterward, sarcasm detection in the text has been extended to a broad range of data forms in social media, such as tweets, comments, and TV dialogues, due to their public availability. Sarcasm detection spanned several approaches like rule-based, supervised, and semi-supervised (\citealp{DBLP2016}) methods, resulting in further development for automatic sarcasm detection. Rule-based methods mainly rely on linguistic information, and their classification accuracy is often not very high due to the presence of noisy data. Most previous work on sarcasm detection based on supervised machine learning tends to rely on different types of features, including sentence length, the number of capitalized words, punctuation~\citep{davidov-etal-2010-semi}, pragmatic factors such as emoticons \citep{gonzalez-ibanez-etal-2011-identifying}, turn-level sentiment lexicon \citep{wilson-etal-2005-recognizing},  sarcasm markers \citep{Ghosh2018}, and so on. Meanwhile, neural models have been applied to this task, relying on semantic relatedness \citep{Amir2016} and neural intra-attention mechanism to capture the sarcasm \citep{Tay2018} and thus reducing feature engineering efforts.

Recently, pre-trained language models such as BERT \citep{Devlin2018}, ERNIE~\citep{sun2019ernie}, and GPT-3~\citep{brown2020language}, have set the new state-of-the-art in a wide range of NLP benchmarks, such as GLUE \citep{GLUE2018}. \citet{Akshay2006} evaluated the performance of pretrained model using feature-based and fine-tuning methods on irony detection in English tweets, finding the latter is better. Meanwhile, there is also a surge of applying pre-trained models in sarcasm detection~\citep{dadu2020sarcasm,potamias2020transformer,javdan2020applying}.
Our system explored the multilingual and monolingual pre-trained language models to testify their fine-tuning performance on English and Arabic sarcasm detection tasks.


 \section{Approach}

\subsection{Pre-trained Language Models}
We adopt \textit{pretrain-then-finetune} paradigm for better leveraging the performance of large-scale pre-trained models. As illustrated in Figure \ref{fig:arch}, for all tasks, we utilize pre-trained models to extract the input representations, followed by a fully-connected feed-forward layer and a softmax/sigmoid activation after the \texttt{[CLS]} token for prediction. For sub-task A and B that input samples only contain one sentence, we directly fine-tune the pre-trained Transformers. For sub-tasks with two sentences, \textit{i.e.}, sub-task C, we employ the multi-layer pre-trained Transformer blocks as the cross-encoder by concatenating sentence pairs and separating them with a \texttt{[SEP]} token.




\begin{figure}[thb]
\centering
\includegraphics[width=\linewidth]{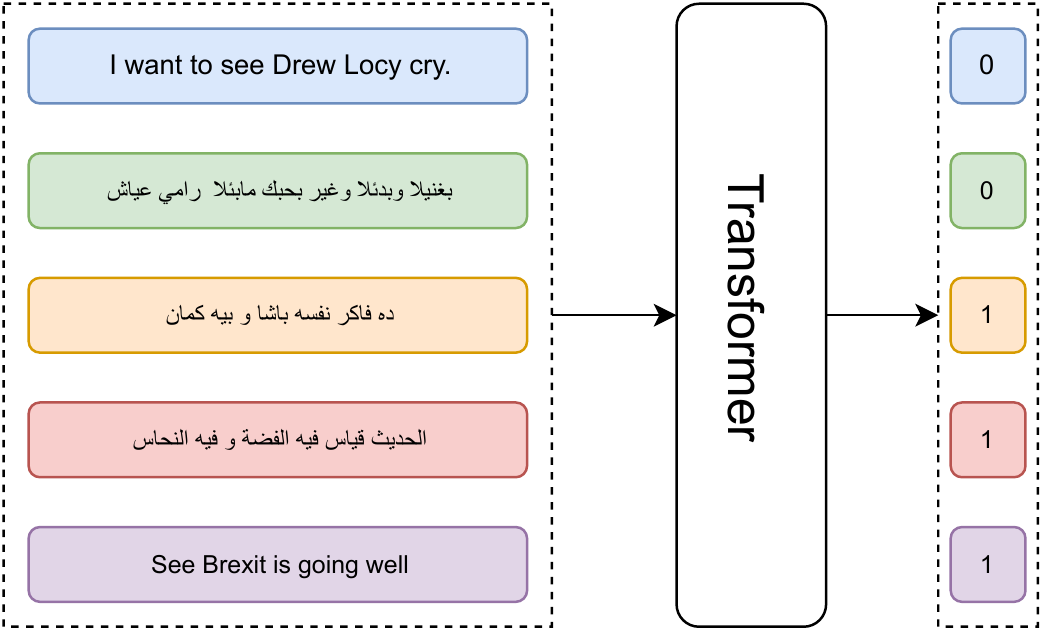}
\caption{Multilingual learning on Task A. ``0/1'' indicate the non-sarcastic and sarcastic class.}
\label{fig:m-a}
\end{figure}

\subsection{Multilingual Learning}
\begin{figure}[t]
\centering
\includegraphics[width=\linewidth]{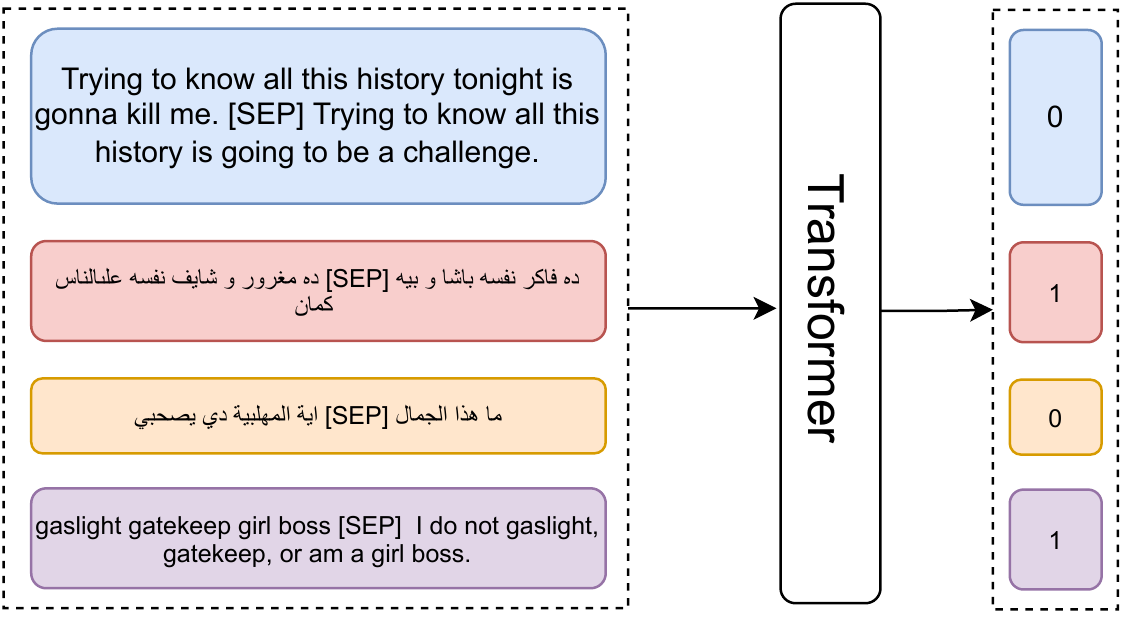}
\caption{Multilingual learning on Task C. ``0/1'' indicate the first or second sentence belongs to sarcasm.}
\label{fig:m-c}
\end{figure}

By observing that subtasks in task A and task C, we found that
both subtasks in Task A and C are for the same objective but in different languages, \emph{i.e.}, Arabic and English. Therefore, we adopt multilingual learning method by simultaneously fine-tuning the pre-trained models on both Arabic and English training data based on multilingual pre-trained models, \emph{i.e.}, ERNIE-M. Specifically, we combine both tasks in Task A or C as a single task, that is, training on Arabic and English sarcasm detection within the same subtask at the same time. As shown in Figure~\ref{fig:m-a}, we combine the one-sentence binary sarcasm detection subtasks in English and Arabic together and fine-tune the multilingual pre-trained models in one forward pass. Similarly, as illustrated in Figure~\ref{fig:m-c}, we conduct the identical settings for Task C. We found that this approach can achieve obvious performance gain on some specific settings and will discuss it in Secion~\ref{sec:res}.

\subsection{Ensemble Learning}
\begin{figure}[thb]
\centering
\includegraphics[width=\linewidth]{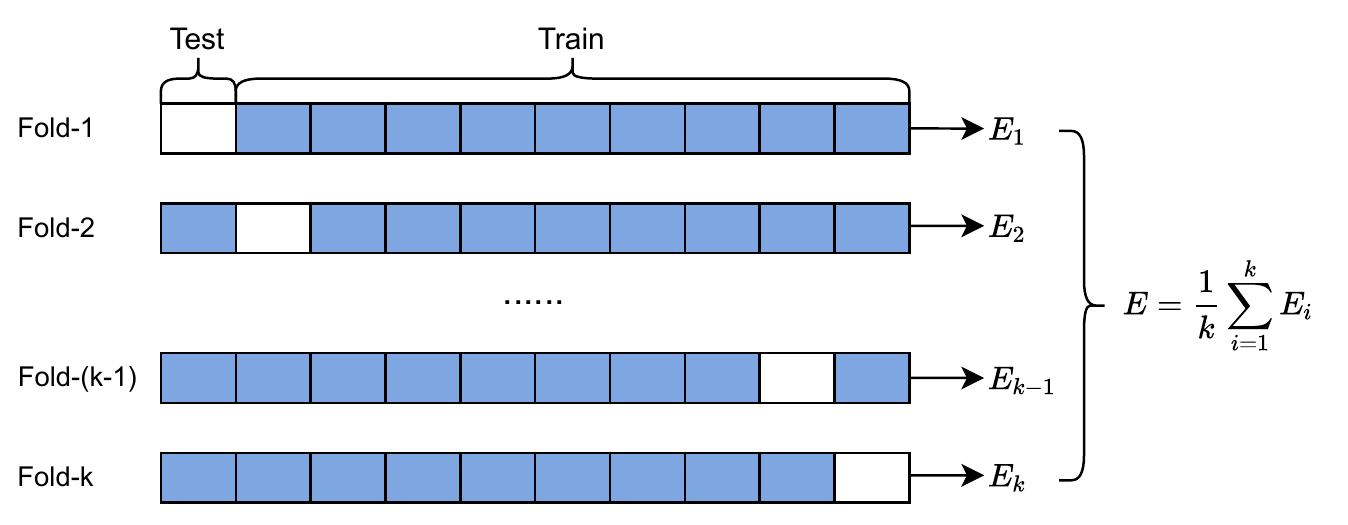}
\caption{Illustration of ensemble learning. $E_i$ indicates the prediction of the $i$-th model on the test set.}
\label{fig:ensemble}
\end{figure}

Considering the limited training data, we split the training data into $k$-fold with disparate random seeds, selecting one out of $k$ data blocks for evaluation and using the rest $k-1$ for data training, as shown in Figure \ref{fig:ensemble}. Then, we choose the optimal model evaluated on various folds and random seeds. Finally, we apply ensemble techniques by averaging all outputs of test sets using optimal models.

\section{Experiments}


\subsection{Task Description}

\subsubsection{Task A: Binary Sarcasm Detection}
The first task is binary text classification: given a tweet sample, the system needs to predict whether it is sarcastic or non-sarcastic. The following examples respectively present a sarcastic and non-sarcastic tweet.

\begin{mylist}
    \item \texttt{The only thing I got from college is a caffeine addiction. \small (\#sarcastic)}
    \item \texttt{ I want to see Drew Lock cry. \small (\#non-sarcastic)}
\end{mylist}

Example 1 is a sarcastic tweet where the author's true intention is "College is really hard, expensive, and exhausting, and I often wonder if the degree is worth the stress."

\subsubsection{Task B: Multi-label Sarcasm Detection}
The second task is a multi-label classification task, where the system requires to predict multiple categories out of six labels, such as \texttt{\#Sarcasm, \#Irony, \#Satire, \#Understatement, \#Overstatement, and \#Rhetorical\_question}. The following examples provide examples for multiple sub-categories:

\begin{mylist}
    \item \texttt{Falling asleep at your laptop is always fun. \small (\#Sarcastic)}
    \item \texttt{Wow Bdubs can bench press 150 kilometers. 	\small (\#Irony)}
    \item \texttt{Lil Pump is the Nelson Mandela of our generation. \small (\#Satire \#Sarcastic)}
    \item \texttt{Lucky for 2nd placed Brentford that there's no stand out team like Leeds this year, or they might have no chance of winning the league. \small (\#Understatement \#Sarcastic)}
    \item \texttt{6 more hours and then a whoppingly massive 2days off work! wowzers! \small (\#Overstatement  \#Irony)}
    \item \texttt{wait what the fuck that solo yolo is mad? \small (\#Rhetorical\_question \#Sarcastic)}
\end{mylist}

In the above examples, the types of sarcasm are subdivided into six categories. \texttt{\#Sarcasm}, which is an ironic remark meant to mock by saying something different than what the speaker really means. For example, in example 1, the speaker hates falling asleep on his laptop.  \texttt{\#Irony} is when something happens that is the opposite of what was expected. As shown in example 2, the fact is that Bdubs cannot bench press 150 kilometers. \texttt{\#Satire} is a type of wit that is meant to mock human vices or mistakes, often through hyperbole, understatement and sarcasm, as shown in Example 3. \texttt{\#Understatement} is often a way of being critical. In example 4, because Norwich is the standout this year, Brentford cannot win the league. \texttt{\#Overstatement} is an act of stating something more profound than it actually is, to make the point more serious, important, or beautiful. In example 5, a whoppingly massive two days off work means regret, and the genuine emotion ought not to require overstatement. \texttt{\#Rhetorical\_question} is a question that is asked even if the person doing the asking knows what the answer is. The solos in example 6 was truly expressed to be awful.

\subsubsection{Task C: Binary Irony Classification on Two Sentences}

The third subtask is binary classification: given a sarcastic tweet and its non-sarcastic rephrase (i.e., two tweets that convey the same meaning), the system needs to predict the sarcastic one. The following examples present a sarcastic sentence and its non-sarcastic paraphrase.

\begin{mylist}
    \item \texttt{Trying to know all this history tonight is gonna kill me. \small (\#Sarcastic)}
    \item \texttt{Trying to know all this history is going to be be a challenge. \small	(\#Rephrase)}
\end{mylist}

\subsection{Evaluation Metrics}

For these three sub-tasks, standard evaluation metrics including accuracy and F1 score are used to evaluate the participating system, calculated as follows:

\begin{small}
\begin{equation}
    accuracy = \frac{TP + TN}{TP+FP+TN+FN} 
\end{equation}
\end{small}

\begin{small}
\begin{equation}
    precision = \frac{TP}{TP+FP}
\end{equation}
\end{small}

\begin{small}
\begin{equation}
recall = \frac{TP}{TP+FN} 
\end{equation}
\end{small}

\begin{small}
\begin{equation}
F_{1} = 2\cdot \frac{precision\cdot recall}{precision+ recall} 
\end{equation}
\end{small}{where $TP, FP, TN, FN$ represent true positive, false positive, true negative, and false negative, respectively.}

As shown in Table \ref{tab:test}, task A, B and C use the F1-score for the sarcastic class, the Macro-F1 score over all classes, and accuracy, respectively. The Macro-F1 score implies that all class labels have equal weights in the final score.

\begin{table}
\centering
\begin{tabular}{lcl}
\hline
\textbf{Task} & \textbf{\#Instances} & \textbf{\#Metric}\\
\hline
\verb|Task A| & {1400}  &{F1-score} \\
\verb|Task B| & {1400} &{Macro-F1 score} \\
\verb|Task C| & {200}  &{accuracy} \\\hline
\end{tabular}
\caption{Summary of official test set in SemEval-2022 Task6.}
\label{tab:test}
\end{table}

\subsection{Data}

The detailed statistics of the sarcasm detection dataset are summarized in Table \ref{tab:t1} and \ref{tab:t2}. As shown in Table \ref{tab:t1}, the training data are shown to be imbalanced, with 867 positive samples vs. 2601 negative ones. We only remove extra spaces, tabs, and line breaks for pre-processing. All emojis that contain emotional factors in training texts are kept without any change. 

\begin{table}
\centering
\begin{tabular}{lc}
\hline
\textbf{Class Label} & \textbf{\#Instances}\\
\hline
\verb|sarcastic| & {867} \\
\verb|non-sarcastic| & {2601} \\
\verb|total| & {3468}  \\\hline
\end{tabular}

\caption{Satirical and non-satirical categories in training data.}
\label{tab:t1}
\end{table}

\begin{table}
\centering
\begin{tabular}{lc}
\hline
\textbf{Multi-class Label} & \textbf{\#Instances}\\
\hline
\verb|sarcasm| & {713} \\
\verb|irony| & {155} \\
\verb|satire| & {25} \\
\verb|understatement| & {10} \\
\verb|overstatement| & {40} \\
\verb|rhetorical_question| & {101}  \\\hline
\end{tabular}

\caption{Six satirical sub-categories in Task B.}
\label{tab:t2}
\end{table}

\subsection{Experiment Details}

Due to the long-tailed nature of training data, we tried to use outside data\footnote{\url{https://www.kaggle.com/c/gse002/data?select=test.csv}} for data augmentation. Particularly, we merge the classes of ``figurative'', ``irony'' in the extra data into a ``sarcasm'' class, but find it of no benefit. We conjecture that this is due to the fact that manual annotators' subjective intention can interpret the same samples with various meanings and therefore result in some noised supervision.

Still, the training data is relatively small and insufficient to achieve an unbiased performance estimate with a random train/test split. Instead, we use a $k$-fold cross-validation procedure ($k=10$), a common model evaluation scheme in machine learning. The $k$-fold cross-validation procedure involves splitting the training dataset into $k$ folds. In which $k-1$ folds are used to train a model, and the rest one fold is used as the evaluation set. Finally, the final output of $k$ models is the mean of these runs.  

For English tasks, we compare ERNIE-M~\citep{ouyang2020ernie} and DeBERTa~\citep{he2021debertav3} as the pre-trained workhorse, while for Arabic tasks, we only consider ERNIE-M. We use the AdamW optimizer \citep{adamw2017} and weight decay of 0.01. We warm up the learning rate for the first 10\% of the update to a peak value of 1e-5 and 5e-6, respectively, and then linearly decay it afterward. We also use dropout \citep{droupout2014} with a rate of 0.15 to prevent overfitting. We adopt a total batch size of 64 by running gradient accumulation on each GPU device with a step size of 8 and a batch size of 1,  sharded across 8 NVIDIA V100 GPU chips. 
Our final solution is to ensemble all the model results obtained using a 10-fold cross-validation strategy with different learning rates (1e-5 and 5e-6) and training epochs (20 and 30), respectively.

\subsection{Results}
\label{sec:res}
Table~\ref{tab:result} compares the final performance on the official test set of task A,B,C under proposed model settings. It is obvious that DeBERTa outperforms ERNIE-M on English task since it is pre-trained only on English corpus. As to the multilingual learning in Task A and C, we observe the significant performance gain (\emph{i.e.}, +6 absolute percentage point on F1 measure) on Task C while find it on par with monolingual fine-tuning on Task A. We guess this is because Task C are given two sentences for comparison, which is more straightforward than Task A (single sentence) to capture the ironic pattern for sarcasm detection. Due to the time limit, we only submit the monolingual fine-tuning results of ERNIE-M (\emph{i.e.}, 84\% acc.), which ranks 5th out of 13 in the Arabic subtask of Task C. Instead, the performance of our multilingual learning can achieve 2nd in Task C (Arabic).
We contend that it would be worthwhile further exploring multilingual learning methods in various language settings in the future.

\begin{table}[]
\resizebox{\linewidth}{!}{%
\begin{tabular}{@{}cccllll@{}}
\toprule
Task & Lang & \multicolumn{1}{c}{\begin{tabular}[c]{@{}c@{}} ERNIE-M \\{\small (multilingual)}\end{tabular}} & \multicolumn{1}{c}{\begin{tabular}[c]{@{}c@{}} ERNIE-M \\{\small (monolingual)}\end{tabular}} & \multicolumn{1}{c}{DeBERTa} & \multicolumn{1}{r}{Rank} \\ \midrule
\multirow{2}{*}{Task A} & en & 36.75 & 38.46 & \textbf{56.91}${({*})}$ & 2/43 \\
 & ar & 40.36 & \textbf{41.87}$({*})$ & - & 9/32 \\ \midrule
Task B & en & N/A & - & \textbf{7.99}$({*})$ & 5/22 \\ \midrule
\multirow{2}{*}{Task C} & en & 82.50 & 75.00 & \textbf{87.00} $({*})$ & 1/16 \\
 & ar & \textbf{90.50} & 84.00${({*})}$ & - & 5/13 \\ \bottomrule
\end{tabular}%
}
\caption{Official test-set performance under various experimental settings. The ``ERNIE-M (multilingual)'' column indicates the performance of multilingual learning in Task A and C. Scores with asterisk indicate final submitted results. The official evaluation metrics for Task A,B,C are F1-score, macro F1-score, and accuracy, respectively.}
\label{tab:result}
\end{table}

\section{Conclusion}

We present our system that participated in SemEval Task 6 and employ the multilingual learning method to train the English and Arabic tasks jointly. We empirically find that it confers benefits in specific scenarios and outranks the monolingual pre-trained models on Arabic tasks. However, we do not adopt other Arabic-specific pre-trained models, which is also worth comparing. In the future, it is a promising direction to explore different sarcasm detection approaches under multilingual settings.



\bibliography{anthology,custom}




\end{document}